\documentclass{article}
\usepackage{iclr2027_conference,times}

\usepackage{amsmath,amsfonts,bm}

\def\eqref#1{equation~\ref{#1}}

\def\1{\bm{1}}

\DeclareMathAlphabet{\mathsfit}{\encodingdefault}{\sfdefault}{m}{sl}
\SetMathAlphabet{\mathsfit}{bold}{\encodingdefault}{\sfdefault}{bx}{n}

\usepackage{hyperref}
\usepackage{url}
\usepackage{graphicx}
\usepackage{wrapfig}
\usepackage{needspace}
\usepackage{xcolor}
\usepackage{listings}
\lstdefinestyle{promptbox}{basicstyle=\scriptsize\ttfamily, frame=single, rulecolor=\color{black!40}, backgroundcolor=\color{black!4}, framesep=4pt, xleftmargin=6pt, xrightmargin=6pt, breaklines=false, keepspaces=true}
\usepackage{booktabs}
\usepackage{multirow}
\usepackage{amsmath}
\usepackage{amssymb}
\usepackage{xcolor}
\usepackage{colortbl}
\makeatletter
\newcommand{\insettablecaption}[1]{\def\@captype{table}\caption{#1}}
\makeatother
\providecommand{\sysname}{GUI-Hopper}

\title{From Tapping to Hopping: Augmenting Mobile GUI Agents with App-Native Deeplinks}

\author{%
\begin{minipage}{0.96\textwidth}
\centering\normalfont\normalsize
Yuchen Sun$^{1,*}$ \quad Chenglin Cai$^{2,*}$ \quad Gongjie Zhang$^{2,*}$\\[3pt]
Tianyu Xia$^{2}$ \quad Quyu Kong$^{2}$ \quad Panrong Tong$^{2}$ \quad Zhengwen Zeng$^{2}$\\[3pt]
Long Chen$^{2}$ \quad Steven Hoi$^{2}$ \quad Chongyang Zhang$^{1,\#}$ \quad Yue Wang$^{2,\#}$\\[5pt]
{\small
$^{1}$Shanghai Jiao Tong University\\[2pt]
$^{2}$Tongyi Lab, Alibaba Group\\[4pt]
$^{*}$Equal contribution. \quad $^{\#}$Corresponding authors.\\[2pt]
\texttt{sunny\_zhang@sjtu.edu.cn} \quad \texttt{wangyue2714@gmail.com}}
\end{minipage}}

\iclrfinalcopy
\hypersetup{%
  pdftitle={From Tapping to Hopping: Augmenting Mobile GUI Agents with App-Native Deeplinks},
  pdfauthor={Yuchen Sun, Chenglin Cai, Gongjie Zhang, Tianyu Xia, Quyu Kong, Panrong Tong, Zhengwen Zeng, Long Chen, Steven Hoi, Chongyang Zhang, Yue Wang},
  pdfsubject={Mobile GUI agents with app-native deeplinks},
  pdfkeywords={Mobile agents, GUI agents, Deeplinks, Hybrid interaction},
  hidelinks
}

\begin{document}

\maketitle

\begin{abstract}
Mobile GUI agents complete tasks using GUI actions like taps and swipes. These actions are broadly applicable across applications, but reaching a task-relevant screen can require many intermediate navigation steps, lengthening trajectories and creating more opportunities for error. Fortunately, many Android applications expose deeplinks---external entry points to specific in-app screens---that provide a higher-level navigation interface. A single deeplink call can replace a sequence of screen-by-screen GUI actions. We therefore introduce hybrid interaction, using deeplinks for direct navigation and GUI actions for other on-screen operations and fallback. To enable this, we discover candidate deeplinks through static analysis, validate them on real devices, and describe their observed landing screens. This process creates a verified and grounded deeplink catalog that pairs each working deeplink with a description of its landing screen. Using this catalog, we introduce \textbf{GUI-Hopper}, a deeplink-augmented GUI agent that learns when to use a deeplink and when to rely on GUI actions. On MobileWorld, GUI-Hopper achieves higher task success with fewer steps than its GUI-only counterpart; its gains persist even without deeplinks at inference. On MobileWorld-Real, it also improves task success in commercial applications on real devices, further demonstrating the benefits of hybrid interaction.
\end{abstract}

\begin{figure}[h!]
\centering
\includegraphics[width=0.95\textwidth]{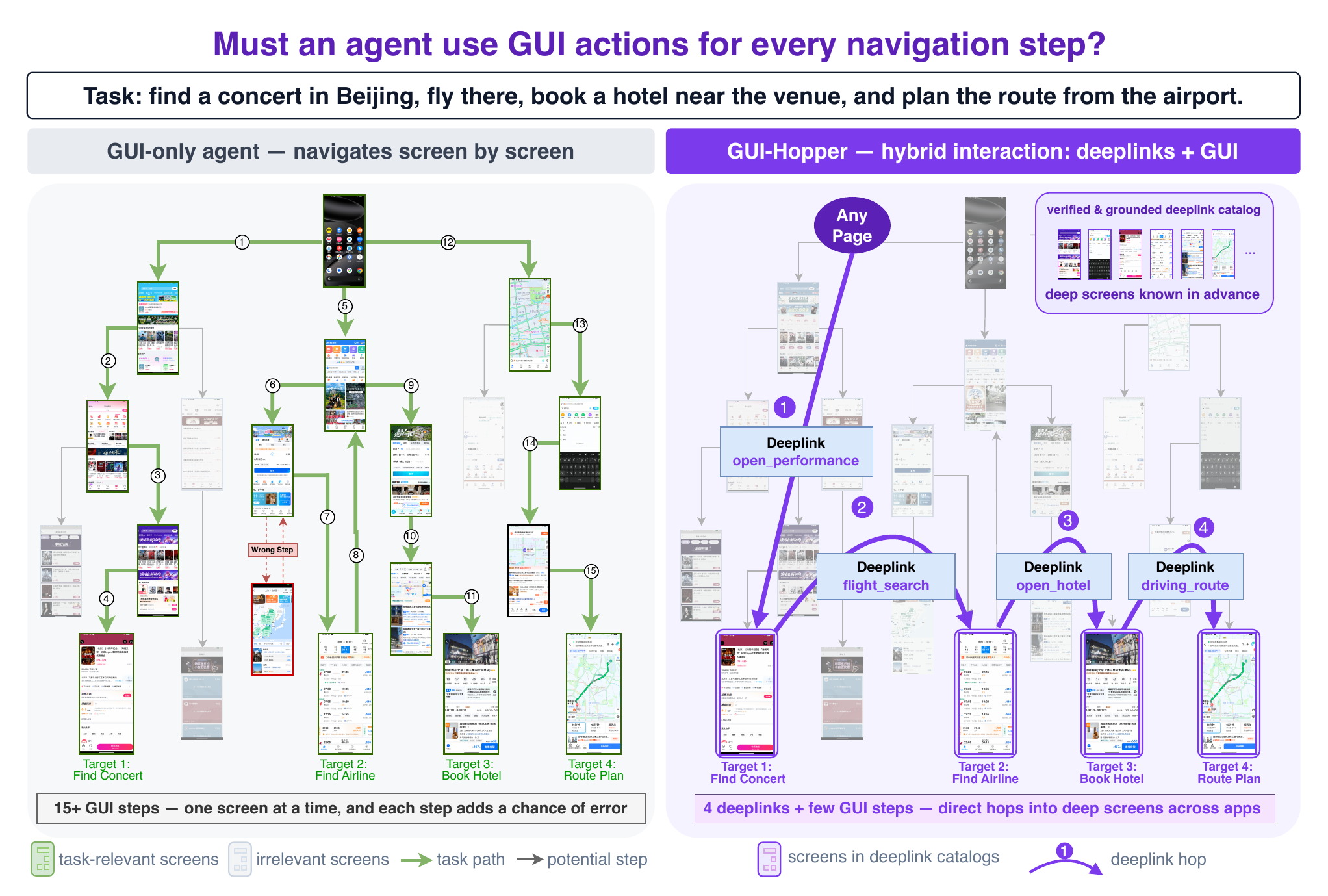}
\vspace{-4mm}
\caption{\textbf{From GUI-only to hybrid interaction.} For a long-horizon mobile task spanning multiple applications, a GUI-only agent navigates screen by screen, requiring 15+ GUI steps (left). GUI-Hopper uses verified deeplinks to reach task-relevant screens directly and GUI actions for the remaining interaction (right), shortening the trajectory and reducing opportunities for error.}
\label{fig:motivation}
\end{figure}

\section{Introduction}

Mobile GUI agents have advanced rapidly with multimodal large language models (MLLMs)~\citep{appagent,uitars,guiowl}. Most operate exclusively through GUI actions: the agent observes the current screen and acts using taps, swipes, etc. This interaction method is native to mobile interfaces, broadly applicable, requires no application-specific API, and supports fine-grained control. Yet relying on GUI actions means that every navigation path must be completed screen by screen. To reach a screen several GUI steps away, a GUI agent must interact with every intermediate screen before the relevant controls become available. Each intermediate screen lengthens the trajectory and introduces another opportunity for error. This raises a practical question: must an agent use GUI actions for every navigation step?

Deeplinks offer an alternative way to navigate. Android applications expose them as external entry points to specific in-app screens~\citep{androidapplinks}. As illustrated by the flight-search example in Figure~\ref{fig:motivation}, a single deeplink call can supply the origin, destination, and travel dates and open the corresponding results screen directly. Beyond reducing navigation steps, a deeplink catalog gives the agent a structured, application-level view of directly reachable destinations and the parameters they accept---information not visible in the current screenshot. Deeplinks and GUI actions therefore play complementary roles: deeplinks provide higher-level navigation, whereas GUI actions support fine-grained on-screen operations and fallback. We call their joint use within a task \emph{hybrid interaction}.

Before an agent can use deeplinks, it must know which ones work and where they lead. We begin by applying existing static-analysis techniques to extract candidate deeplinks~\citep{masbench}. However, these candidates are not yet suitable for use as agent actions: some are stale or redundant, some require parameter values that are not specified in their declarations, and their behavior may depend on runtime context. Moreover, a deeplink URI alone rarely conveys what the deeplink does or where it leads. To resolve this ambiguity, we develop an automated pipeline that searches for valid parameter values when needed and directly executes each candidate on real devices. The pipeline discards candidates that fail or do not lead to a meaningful screen and pairs each retained deeplink with a language description of its observed landing screen. Together, these steps transform raw candidates into a \emph{verified and grounded deeplink catalog}.

A reliable catalog alone is not enough: an agent must still determine whether a given deeplink will help with the current task. We therefore introduce \textbf{GUI-Hopper}, a deeplink-augmented GUI agent trained to use a deeplink when it is useful and otherwise rely on GUI control. To teach this decision, negative augmentation exposes GUI-Hopper to cases in which the matching deeplink is withheld, only irrelevant deeplinks are provided, or an invocation fails at runtime. We then optimize GUI-Hopper through online reinforcement learning to encourage correct deeplink use and efficient task completion. At inference, the system retrieves deeplinks whose landing-screen descriptions match the task, and GUI-Hopper decides whether to invoke one. After each invocation, the system checks whether the expected screen has been reached. If no suitable deeplink is available or verification fails, GUI-Hopper continues the task through ordinary GUI interaction.

We evaluate whether this combination improves task completion and interaction efficiency. On MobileWorld~\citep{mobileworld}, GUI-Hopper completes more tasks with fewer interaction steps than a matched GUI-only agent, and the advantage holds across the three model scales we train. GUI-Hopper continues to outperform the GUI-only agent when evaluated without deeplinks, showing that its gains persist beyond direct deeplink use. GUI-Hopper also performs well on real devices in MobileWorld-Real~\citep{qwenuiagent}. Together, these results point to a paradigm shift: from GUI-only mobile agents to agents that combine GUI control with deeplinks when they are available.

The contributions of this paper are threefold. 
First, we introduce GUI-Hopper, which combines deeplinks for higher-level navigation with GUI actions for fine-grained on-screen operations and fallback. This allows the agent to reach task-relevant screens directly when suitable deeplinks are available while retaining GUI control throughout the task. To the best of our knowledge, we are the first to systematically integrate deeplinks into mobile GUI agents across catalog construction, model training, and inference-time execution.
Second, we develop an automatic deeplink mining and grounding pipeline that discovers candidate deeplinks, tests them on real devices, and pairs each verified deeplink with a description of its observed landing screen. Third, we train GUI-Hopper to select deeplinks and return to GUI control when needed, using negative augmentation and reinforcement learning with rewards for task success, deeplink use, and interaction efficiency. Experiments show that GUI-Hopper improves task success and interaction efficiency, retains its gains when evaluated without deeplinks, and performs well on real devices.

\section{Related Work}

\paragraph{Mobile GUI Agents.}
Existing work on mobile GUI agents can be broadly divided into two categories. Framework-based approaches orchestrate off-the-shelf multimodal models through prompting, ranging from single-agent exploration to multi-agent collaboration with memory \citep{appagent, mobileagent, mobileagente}. More recent approaches instead train native GUI models end to end, scaling unified architectures across platforms through data flywheels and online reinforcement learning \citep{uitars, uitars2, maiui, uivenus15, guiowl, stepgui, xiaomigui}. 
Recent benchmarks emphasize longer tasks, cross-application workflows, and real-device interaction~\citep{mobileworld,androiddaily,qwenuiagent}, making trajectory efficiency important alongside task success~\citep{odysseys,androidlab}.
Yet despite these advances, agents in both categories still perform every navigation step through the GUI, screen by screen---the inefficiency targeted by our work.

\paragraph{Tool-Augmented Mobile Agents.}

Tool-augmented mobile agents complement GUI interaction with tools that execute reusable action sequences or directly invoke functions within applications.
Academic approaches generally construct tools on the agent side, either by distilling reusable skills from agent trajectories \citep{appagentx, mobileagente} or by building tool libraries from documentation and synthetic traces; several studies further train models to invoke these tools \citep{ultracua, toolcua, droidcall}. These agent-constructed tools can be brittle: trajectory-derived skills may be tied to the UI versions on which their source trajectories were collected, while synthesized tool libraries may not behave reliably in real applications.
Industrial systems, by contrast, rely on platform interfaces such as Apple App Intents and Google App Actions that application developers integrate directly \citep{appintents, appactions}. These interfaces are dependable once integrated, but their coverage is limited by developer adoption.  
Deeplinks offer a third path that avoids much of this trade-off. Because they are app-native, deeplinks can be reused without requiring developers to implement a new agent-facing API. When mined systematically, they also expose an application's directly addressable pages. Recent work provides agents with manually curated deeplinks at inference time~\citep{masbench,knowact}. We go further by automatically mining and validating deeplinks on real devices, training the agent to choose between deeplinks and GUI actions, and supporting inference with retrieval and landing-screen verification. To our knowledge, no prior work addresses this full pipeline.

\paragraph{Deeplinks in the Android Ecosystem.}
Android applications expose deeplinks as external entry points to directly launch specific in-app screens~\citep{androidapplinks}.
Research on deeplinks has examined their generation, use in software testing, and security. On the generation side, uLink and Aladdin generate deeplinks for existing applications, motivated by the observation that many applications expose few developer-defined links \citep{ulink, aladdin}. In software testing, deeplinks improve the coverage of automated exploration by providing direct access to deeply nested pages \citep{delm}. Security studies have measured hijacking risks at scale and shown that the verification machinery can itself be bypassed \citep{liu2017deeplink, tang2020applinks}. Recent analyses further report that production OEM assistants already invoke deeplinks to bypass screen-by-screen UI navigation \citep{wu2025assistants}. 
Collectively, this literature establishes deeplinks as a mature mechanism in Android. It does not, however, study how a trained mobile agent can learn to combine deeplinks with GUI actions.

\section{Method}

\begin{figure}[t!]
\vspace{-5pt}
\centering
\includegraphics[width=0.999\textwidth]{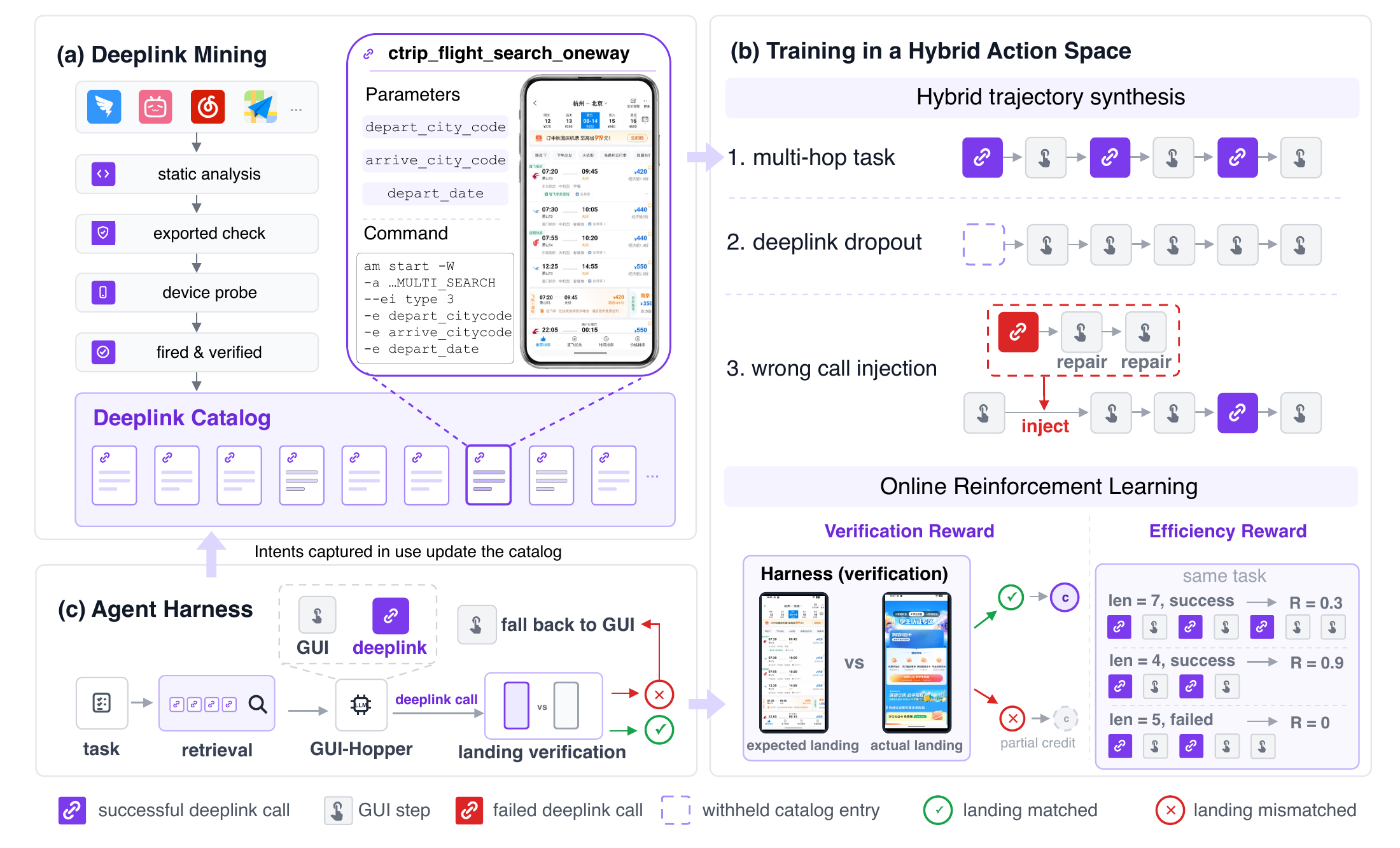}
\vspace{-20pt}
\caption{\textbf{Overview of our approach.} (a) A mining pipeline verifies deeplinks on real devices and builds a grounded catalog. (b) Hybrid trajectory synthesis and online reinforcement learning train GUI-Hopper in the hybrid action space. (c) At inference, the harness retrieves task-specific deeplink candidates and checks whether each call reaches its expected Activity, with GUI fallback on failure.}
\label{fig:method}
\end{figure}

\sysname{} is a mobile agent trained to use deeplink actions alongside GUI control. GUI-Hopper is built on three key components (Figure~\ref{fig:method}): a mining pipeline that turns app-provided deeplinks into a verified and grounded catalog (Section~\ref{sec:mining}), a hybrid-action training strategy that teaches the model when to invoke a deeplink and when to continue with GUI actions (Section~\ref{sec:training}), and an agent harness for verified deeplink use on real phones (Section~\ref{sec:harness}).

\subsection{Preliminaries}\label{sec:prelim}

\paragraph{Deeplinks in Android.}
Android applications organize user-facing screens into \emph{Activities}. An Activity is launched through an \emph{Intent}, a system message that identifies the target component and carries invocation parameters. In \texttt{AndroidManifest.xml}, an application can associate URI patterns with Activities that external callers may open. Android resolves an Intent carrying the URI to an Activity that can handle it and launches that Activity; we refer to this URI-based entry point as a deeplink~\citep{androidintents}. Because a raw URI rarely reveals its landing screen, we expose each deeplink to the model as a tool with a name, a landing-screen description, and parameter definitions. Section~\ref{sec:mining} describes how these fields are constructed. In the formulation below, a deeplink action invokes one such catalog entry with concrete parameters.

\paragraph{Task formulation.}
We model mobile interaction as a partially observable Markov decision process~\citep{kaelbling1998pomdp} with environment states \(s_t \in \mathcal{S}\), observations \(o_t \in \mathcal{O}\), actions \(a_t \in \mathcal{A}\), and transition dynamics \(\mathcal{T}(s_{t+1} \mid s_t, a_t)\). The action space is hybrid: \(\mathcal{A} = \mathcal{A}_{\mathrm{GUI}} \cup \mathcal{A}_{\mathrm{DL}}\). Here, \(\mathcal{A}_{\mathrm{GUI}}\) contains atomic screen operations such as tapping, swiping, and typing. Each deeplink action is a pair \((d, x)\), where \(d\) is a catalog entry supplied for the current task and \(x\) assigns concrete values to its parameters. At step \(t\), the agent receives an observation \(o_t = (v_t, f_t)\), where \(v_t\) is the current screenshot and \(f_t\) is the harness's feedback after the preceding action. GUI actions produce no structured feedback. For a deeplink action, \(f_t\) reports whether the invocation succeeded and whether the observed landing Activity matches the expected Activity recorded in the catalog. The policy chooses \(a_t = \pi_\theta(\mathcal{I}, o_t, \mathcal{H}_t)\) from the task instruction \(\mathcal{I}\), the current observation \(o_t\), and the recent interaction history \(\mathcal{H}_t = (o_{t-h}, a_{t-h}, \ldots, o_{t-1}, a_{t-1})\). We optimize the policy parameters to maximize the expected trajectory reward \(J(\theta) = \mathbb{E}_{\tau \sim \pi_\theta}[\mathcal{R}(\tau)]\), with \(\mathcal{R}\) defined in Section~\ref{sec:training}.

\subsection{Deeplink Mining and Description}\label{sec:mining}

Android applications already expose deeplink entry points, but their declarations alone do not make these entries usable as agent actions. Building the catalog requires two steps: verifying which candidates work on a device and representing each verified entry with a landing-screen description and parameter definitions. Static analysis provides initial candidates but cannot determine whether a candidate works or where it lands. This limitation is especially important for entries that route to different screens depending on parameter values not specified in their declarations. We therefore test candidates on real devices and use the observed results to verify and describe each retained entry.

\paragraph{Mining pipeline.}
The pipeline (Figure~\ref{fig:method}(a)) begins by extracting candidate entries from manifest declarations and shortcut definitions~\citep{androidintents,androidshortcuts} through static analysis of the application package. We first filter out candidates whose target components are not exported and therefore cannot be invoked externally. For the remaining candidates, we use a read-only package-manager query to check whether they resolve to installed components, without launching the application or invoking a model. Finally, we test each resolvable candidate on a real device and retain only those that open a meaningful screen. For parameterized entries, we test candidate values and retain a parameter only if changing its value visibly changes the resulting screen. After verification, we consolidate duplicate entries and exclude those pointing to transient targets to form the final catalog. Across the 26 commercial applications selected for MobileWorld-Real evaluation, the resulting catalog contains 1,251 verified entries.

\paragraph{Describing deeplinks for agent use.}
For each verified deeplink, we create a catalog entry describing where it leads and how to invoke it. Each entry specifies a tool name, a description of the observed landing screen, and parameter definitions with formats and example values. Since a screen often supports multiple functions, we describe the landing screen rather than a single function. We also specify what remains to be done after landing, making clear that reaching the screen does not necessarily mean that the task is complete. Together, these details help a mobile agent select a task-relevant deeplink and supply appropriate parameter values. To enable the harness to execute the call and verify its landing, each entry also stores an invocation template and the expected landing Activity. The agent harness fills this template with the selected parameter values to construct the call, then compares the observed Activity with the expected one. Appendix~\ref{app:schema} presents the complete catalog entry format and prompt format, while Appendix~\ref{app:mining} provides implementation details and statistics.

\subsection{Training in a Hybrid Action Space}\label{sec:training}

Training in a hybrid action space requires more than learning the syntax of a deeplink call. The model must decide whether to invoke a deeplink or continue with GUI actions as the available deeplink set varies, and it must recover when a call fails. We first use hybrid trajectory synthesis and negative augmentation to teach selection and recovery. We then apply online reinforcement learning to refine these behaviors.

\paragraph{Hybrid trajectory synthesis.}
We synthesize training trajectories from tasks constructed around the mined catalog (Figure~\ref{fig:method}(b)). They cover both tasks that chain multiple deeplink calls and tasks that require GUI actions after a deeplink call, demonstrating how the two action types can be combined. Because an unsuitable deeplink may take the agent directly to an unrelated screen, our negative augmentation targets both prevention and recovery. To prevent incorrect calls, we introduce \emph{deeplink dropout}: we sample successful deeplink trajectories and collect GUI-only trajectories for the same tasks, supplying only task-irrelevant deeplinks in the prompt. These trajectories demonstrate how to fall back on GUI actions when retrieval provides no useful deeplinks. To train recovery, we apply \emph{wrong-call injection} during trajectory generation. We deliberately invoke an unsuitable deeplink and retain the subsequent sequence in which the agent recognizes that the landing screen is irrelevant to the task, recovers from the error, and resumes the task. Together, these trajectories teach the agent to use suitable deeplinks, rely on GUI actions when none are suitable, and recover from incorrect calls.

\paragraph{Online reinforcement learning.}
The synthesized trajectories teach the model the basic decision pattern, while online interaction provides direct feedback on whether its deeplink choices reach the expected destination and contribute to successful, efficient task completion. We therefore further optimize \sysname{} with online reinforcement learning using both call-level and trajectory-level feedback. Each catalog entry records its verified landing Activity. After execution, the feedback $f_t$ reports whether the invocation succeeded and whether the observed Activity matches the expected one (Section~\ref{sec:prelim}), providing a call-level verification signal. At the trajectory level, $s(\tau)$ records task success, while $|\tau|$ counts the number of interaction steps and provides the basis for the efficiency term. We derive the trajectory reward from these quantities:
\begin{equation}
\mathcal{R}(\tau) = s(\tau)\,\bigl(1 + \alpha\, c(\tau) + \beta\, e(\tau)\bigr),
\end{equation}
where $s(\tau) \in \{0,1\}$ indicates whether the task is completed. The call score $c(\tau)$ evaluates each task-relevant deeplink, assigning full credit when it is invoked and reaches its expected landing, partial credit when it is invoked but fails the landing check, and zero when it is not invoked. The efficiency score $e(\tau) = \max(0,\,1-|\tau|/ S_{\mathrm{ref}})$ is zero for trajectories at or above the reference length $S_{\mathrm{ref}}$ and increases linearly as trajectories become shorter. Because $s(\tau)$ gates the entire reward, a failed trajectory receives zero reward regardless of its deeplink calls or length. Among successful trajectories, the reward increases with the call score and with reductions in length below $S_{\mathrm{ref}}$. We compute advantages with the group relative policy optimization (GRPO) estimator~\citep{deepseekmath}. All trajectories within a GRPO group receive the same deeplink candidate set, so their relative advantages are computed under the same action space. Candidate sets vary across groups to expose the model to different deeplink availability without mixing conditions within a group.

\subsection{Agent Harness}\label{sec:harness}

The mining pipeline in Section~\ref{sec:mining} produces a verified and grounded deeplink catalog, while the training strategy in Section~\ref{sec:training} teaches the model to choose between deeplink and GUI actions. 
To use the catalog and trained model together on real phones, the agent harness retrieves task-specific deeplinks, executes and verifies deeplinks with GUI fallback, and captures parameter values observed during use (Figure~\ref{fig:method}(c)).

\paragraph{Task-level retrieval.}
Exposing the full catalog at every step would consume substantial context and present the model with a large choice set. We therefore construct a task-specific candidate set before execution. The model decomposes the task into ordered navigation goals and identifies the application associated with each goal. For each goal, the harness restricts the candidate pool to that application and ranks the entries using semantic and lexical matching. It then merges the per-goal rankings and returns at most 10 deeplinks. The agent uses this compact candidate set throughout the task instead of the full catalog.

\paragraph{Landing verification and fallback.}
A deeplink is invoked through a system command, whose return status alone does not establish whether the intended screen was reached. With incorrect parameter values, the application may open an unexpected screen without reporting an error. Our harness therefore checks whether the call executed successfully and whether the observed landing Activity matches the expected one recorded in the catalog. It returns both results as the feedback \(f_t\) defined in Section~\ref{sec:prelim}. If either check fails, the next observation makes the failure explicit, allowing the agent to recover through GUI actions rather than continue as though the deeplink had succeeded.

\paragraph{Deeplink parameter capture during use.}
Deeplink parameters such as dates, locations, and item identifiers allow later calls to reuse information from a user's interaction history. During ordinary app use, the harness observes the Intents issued by the application, extracts reusable parameter values, and associates them with the corresponding catalog entries. The agent can then invoke these entries with values observed from that user, for example to reopen the page for a previously booked hotel or retrieve a past order by its identifier.

Together, the three components implement the deeplink-specific parts of the formulation in Section~\ref{sec:prelim}: retrieval constructs a task-specific candidate set from \(\mathcal{A}_{\mathrm{DL}}\), verification produces \(f_t\), and parameter capture adds observed values \(x\) for later calls \((d,x)\). Additional implementation details are provided in Appendix~\ref{app:harness}.

\section{Experiments}

\subsection{Setup}\label{sec:setup}

\paragraph{Benchmarks and metrics.}
We evaluate \sysname{} on MobileWorld~\citep{mobileworld} and MobileWorld-Real~\citep{qwenuiagent}. MobileWorld provides a simulated Android environment with 13 built-in applications, where we evaluate all 117 tasks. We evaluate each MobileWorld task three times per setting and compute success rate over all 351 task executions. For MobileWorld-Real, we select 26 frequently used applications from the full benchmark and apply the proposed automated pipeline to mine their deeplinks. We then select 119 tasks involving these applications from the full benchmark for evaluation. For both benchmarks, we report task success rate and the average number of interaction steps for successful trajectories and for all trajectories.

\begin{table}[!t]
\centering
\setlength{\abovecaptionskip}{0pt}
\caption{Experiment results on MobileWorld. GUI-Hopper improves success and efficiency over GUI-only training across three model scales. Its success-rate gains persist under GUI-only evaluation.}
\label{tab:mw-main}
\begin{minipage}{0.90\linewidth}
\centering
\fontsize{9}{10.5}\selectfont
\setlength{\tabcolsep}{3pt}
\setlength{\aboverulesep}{1pt}
\setlength{\belowrulesep}{1pt}
\renewcommand{\arraystretch}{1.0}
\begin{tabular*}{\linewidth}{@{\extracolsep{\fill}}lcccc@{}}
\toprule
\multirow{2}{*}{\makebox[0.49\linewidth][l]{\textbf{Model}}} & \multirow{2}{*}{\shortstack{\textbf{Eval. with}\\\textbf{deeplink}}} & \multirow{2}{*}{\textbf{SR (\%)} $\uparrow$} & \multicolumn{2}{c}{\textbf{Avg. length (steps)} $\downarrow$} \\
 & & & \mbox{Success-only} & \makebox[30pt]{All} \\
\midrule
\rowcolor[HTML]{F1F3F5}
\multicolumn{5}{c}{\textbf{General-purpose VLMs}} \\
Claude Opus 4.6~\citep{claudeopus46} & $\times$ & 44.4 & 18.2 & 28.5 \\
Gemini 3.1 Pro~\citep{gemini31pro} & $\times$ & 58.1 & 21.1 & 25.6 \\
GPT-5.6-Sol~\citep{gpt56} & $\times$ & 70.1 & 21.9 & 26.5 \\
Seed-2.0-Pro~\citep{seed20} & $\times$ & 63.3 & 21.3 & 26.9 \\
\midrule
\rowcolor[HTML]{F1F3F5}
\multicolumn{5}{c}{\textbf{Specialized GUI Models}} \\
GUI-Owl-1.5-4B-Instruct~\citep{guiowl} & $\times$ & 25.1 & 16.1 & 28.4 \\
GELab-Zero-4B~\citep{stepgui} & $\times$ & 16.0 & 14.2 & 29.9 \\
PhoneBuddy-4B~\citep{phonebuddy} & $\times$ & 23.1 & 12.1 & 27.8 \\
ForgeOwl-8B~\citep{mobileforge} & $\times$ & 41.3 & 16.8 & 28.2 \\
AutoGLM-Phone-9B~\citep{autoglm} & $\times$ & 15.0 & 18.6 & 35.0 \\
UI-TARS-1.5-7B~\citep{uitars15} & $\times$ & 8.9 & 13.1 & 28.4 \\
MAI-UI-8B~\citep{maiui} & $\times$ & 27.6 & 17.8 & 36.8 \\
UI-Venus-1.5-30B-A3B~\citep{uivenus15} & $\times$ & 19.9 & 12.3 & 31.0 \\
UI-Venus-2-9B~\citep{uivenus2} & $\times$ & 68.1 & 19.9 & 24.7 \\
\midrule
\rowcolor[HTML]{EFE7FD}
\multicolumn{5}{c}{\textbf{Qwen3.5-based models}} \\
\textbf{Qwen3.5-4B} & $\times$ & 15.7 & 12.4 & 32.9 \\
\hspace{0.7em}+ GUI training & $\times$ & 52.7 & 19.2 & 26.6 \\
\hspace{0.7em}+ Hybrid training (ours) & $\times$ & 53.0 & 18.9 & 23.5 \\
\hspace{0.7em}+ Hybrid training (ours) & $\checkmark$ & \textbf{60.1} & 17.5 & 20.9 \\
\midrule[0.3pt]
\textbf{Qwen3.5-9B} & $\times$ & 20.8 & 16.5 & 30.2 \\
\hspace{0.7em}+ GUI training & $\times$ & 54.4 & 19.4 & 25.7 \\
\hspace{0.7em}+ Hybrid training (ours) & $\times$ & 56.7 & 19.1 & 23.1 \\
\hspace{0.7em}+ Hybrid training (ours) & $\checkmark$ & \textbf{63.5} & 17.5 & 21.8 \\
\midrule[0.3pt]
\textbf{Qwen3.5-35B-A3B} & $\times$ & 17.7 & 13.0 & 27.5 \\
\hspace{0.7em}+ GUI training & $\times$ & 57.8 & 19.2 & 25.2 \\
\hspace{0.7em}+ Hybrid training (ours) & $\times$ & 61.8 & 19.0 & 24.3 \\
\hspace{0.7em}+ Hybrid training (ours) & $\checkmark$ & \textbf{64.1} & 16.7 & 21.0 \\
\bottomrule
\end{tabular*}
\end{minipage}
\end{table}

\begin{table}[!t]
\centering
\setlength{\abovecaptionskip}{0pt}
\caption{Experiment results on MobileWorld-Real, a real-device mobile GUI agent benchmark. GUI-Hopper improves success and efficiency over GUI-only training on physical devices.}
\label{tab:real-main}
\begin{minipage}{0.90\linewidth}
\centering
\fontsize{9}{10.5}\selectfont
\setlength{\tabcolsep}{3pt}
\setlength{\aboverulesep}{1pt}
\setlength{\belowrulesep}{1pt}
\renewcommand{\arraystretch}{1.0}
\begin{tabular*}{\linewidth}{@{\extracolsep{\fill}}lcccc@{}}
\toprule
\multirow{2}{*}{\makebox[0.49\linewidth][l]{\textbf{Model}}} & \multirow{2}{*}{\shortstack{\textbf{Eval. with}\\\textbf{deeplink}}} & \multirow{2}{*}{\textbf{SR (\%)} $\uparrow$} & \multicolumn{2}{c}{\textbf{Avg. length (steps)} $\downarrow$} \\
 & & & \mbox{Success-only} & \makebox[30pt]{All} \\
\midrule
\rowcolor[HTML]{F1F3F5}
\multicolumn{5}{c}{\textbf{General-purpose VLMs}} \\
Claude Opus 4.6~\citep{claudeopus46} & $\times$ & 75.0 & 21.8 & 25.9 \\
Gemini 3.1 Pro~\citep{gemini31pro} & $\times$ & 96.9 & 14.9 & 15.6 \\
GPT-5.6-Sol~\citep{gpt56} & $\times$ & 96.6 & 21.2 & 21.7 \\
Kimi K2.6~\citep{kimik26} & $\times$ & 68.9 & 19.8 & 28.6 \\
Qwen3.7-Plus~\citep{qwen37plus} & $\times$ & 74.7 & 20.0 & 27.1 \\
Seed-2.1-Pro~\citep{seed21} & $\times$ & 89.6 & 20.2 & 22.8 \\
\midrule
\rowcolor[HTML]{F1F3F5}
\multicolumn{5}{c}{\textbf{Specialized GUI Models}} \\
GELab-Zero-4B~\citep{stepgui} & $\times$ & 34.1 & 20.5 & 35.3 \\
GUI-Owl-1.5-32B-Instruct~\citep{guiowl} & $\times$ & 10.0 & 23.7 & 40.4 \\
UI-Venus-1.5-30B-A3B~\citep{uivenus15} & $\times$ & 43.8 & 11.3 & 31.9 \\
UI-Venus-2-9B~\citep{uivenus2} & $\times$ & 84.2 & 17.7 & 22.4 \\
Qwen-UI-Agent-27B~\citep{qwenuiagent} & $\times$ & 92.0 & 16.5 & 18.1 \\
\midrule
\rowcolor[HTML]{EFE7FD}
\multicolumn{5}{c}{\textbf{Qwen3.5-based models}} \\
\textbf{Qwen3.5-4B} & $\times$ & 21.1 & 13.7 & 33.4 \\
\hspace{0.7em}+ GUI training & $\times$ & 70.2 & 17.6 & 26.5 \\
\hspace{0.7em}+ Hybrid training (ours) & $\checkmark$ & \textbf{84.0} & 16.7 & 20.4 \\
\midrule[0.3pt]
\textbf{Qwen3.5-9B} & $\times$ & 18.9 & 14.4 & 30.6 \\
\hspace{0.7em}+ GUI training & $\times$ & 77.1 & 17.4 & 24.2 \\
\hspace{0.7em}+ Hybrid training (ours) & $\checkmark$ & \textbf{86.6} & 16.6 & 20.4 \\
\midrule[0.3pt]
\textbf{Qwen3.5-35B-A3B} & $\times$ & 39.5 & 12.8 & 32.1 \\
\hspace{0.7em}+ GUI training & $\times$ & 79.2 & 17.3 & 23.7 \\
\hspace{0.7em}+ Hybrid training (ours) & $\checkmark$ & \textbf{89.1} & 15.7 & 18.2 \\
\bottomrule
\end{tabular*}
\end{minipage}
\end{table}

\paragraph{Training settings.}
For MobileWorld, we use Qwen3.5~\citep{qwen35} models at three scales: 4B, 9B, and 35B-A3B. At each scale, we compare the base model, an agent trained with GUI actions alone, and GUI-Hopper. The two trained agents use paired trajectories from the same tasks and otherwise follow identical training configurations; they differ only in whether deeplink actions and their associated feedback are available. We perform supervised fine-tuning for six epochs with a peak learning rate of $3\times10^{-6}$. After a warmup over the first 5\% of training steps, the learning rate follows a cosine schedule and decays to $1\times10^{-7}$. We use a global batch size of 64, a microbatch size of 1, and a context length of 28,672 tokens. Online reinforcement learning uses a group size of 8, a rollout temperature of 1.0, and a constant learning rate of $1\times10^{-6}$, with $\alpha=0.15$, $\beta=0.10$, and $S_{\mathrm{ref}}=25$. Appendix~\ref{app:training} provides the complete training configuration.

\paragraph{Construction of deeplink catalogs.}
We construct the deeplink catalogs using the mining, verification, and grounding pipeline described in Section~\ref{sec:mining}. The pipeline yields 34 entries for MobileWorld's built-in applications and 1,251 entries for the 26 selected MobileWorld-Real applications. At inference, we retrieve up to 10 candidate deeplinks from the catalog based on the current task and provide them to GUI-Hopper. In the main comparisons, the base model and GUI-only agent receive no deeplink candidates and operate solely in $\mathcal{A}_{\mathrm{GUI}}$.

\subsection{Main Results}

\subsubsection{MobileWorld}

Table~\ref{tab:mw-main} compares the base model, the GUI-only agent, and GUI-Hopper across three model scales. Both trained agents substantially improve over their corresponding base models. Across all three model scales, GUI-Hopper with access to deeplinks consistently achieves higher success rates than the GUI-only agent using GUI actions alone. It reduces average steps across all trajectories by 15.26--21.30\%, with shorter successful trajectories at every scale. Verified deeplinks can take the agent directly to task-relevant screens, bypassing lengthy GUI navigation and reducing opportunities for navigation errors in long-horizon tasks.

Even when evaluated with GUI actions alone, GUI-Hopper's hybrid-SFT checkpoints achieve higher success rates than the GUI-only agent across all three model scales. GUI-Hopper also completes tasks in fewer GUI steps than the GUI-only agent. Together, these results show that the hybrid-trained model retains effective GUI ability: its gains are not confined to executions that can use direct navigation shortcuts. Appendix~\ref{sec:trajectory-length} provides a visualization of the successful-trajectory length distributions at 9B.

One possible explanation is that hybrid-action training encourages the model to break tasks into intermediate goals and identify the screens needed to achieve them. This pattern, learned with hybrid interactions, may also generalize to mobile use with solely GUI actions, helping the agent avoid unnecessary navigation. An illustrative GUI-only case is provided in the appendix (Figure~\ref{fig:app-case-gui-planning}).

\subsubsection{MobileWorld-Real}

We deploy the complete pipeline on MobileWorld-Real to verify whether hybrid interaction generalizes to commercial applications running on physical devices. As summarized in Table~\ref{tab:real-main}, GUI-Hopper consistently outperforms the GUI-only agent in both task completion and trajectory efficiency, confirming that the efficiency advantages of deeplink-guided navigation effectively translate from simulation to complex real-world mobile environments.

\subsection{Ablations}

\subsubsection{Contribution of Training Stages}

\begin{wraptable}{r}{0.49\textwidth}
\vspace{\dimexpr-\intextsep-1.8\baselineskip\relax}
\setlength{\abovecaptionskip}{0pt}
\caption{Training-stage ablation on Qwen3.5-4B. Online RL further improves success and efficiency after hybrid SFT.}
\label{tab:abl-training}
\centering
\fontsize{8}{9.5}\selectfont
\setlength{\tabcolsep}{3pt}
\renewcommand{\arraystretch}{1.0}
\begin{tabular*}{\linewidth}{@{\extracolsep{\fill}}ccc@{}}
\toprule
Stage & SR (\%) $\uparrow$ & Length $\downarrow$ \\
\midrule
Base (GUI) & 15.7 & 32.9 \\
+ Hybrid SFT & 57.8 & 22.5 \\
+ Online RL & \textbf{60.1} & \textbf{20.9} \\
\bottomrule
\end{tabular*}
\par\smallskip
\raggedright Trained variants use hybrid evaluation. Length: average steps over all trajectories.\par
\vspace{-\intextsep}
\end{wraptable}

We examine the contribution of the training stages (Table~\ref{tab:abl-training}). Both the hybrid-SFT model and its online-RL refinement use Deeplink-Ten at evaluation. We include the base model evaluated with GUI actions as a reference.

Online reinforcement learning increases success rate from 57.83\% to 60.11\%, while reducing the average number of steps across all trajectories from 22.52 to 20.91. With the deeplink supply unchanged, these gains suggest that online feedback helps the agent make better use of the hybrid action space learned during supervised training.

\vspace{-1pt}
\subsubsection{Effect of Deeplink Supply}

\begin{wraptable}{r}{0.52\textwidth}
\vspace{\dimexpr-\intextsep-1.8\baselineskip\relax}
\vspace{20pt}
\setlength{\abovecaptionskip}{0pt}
\caption{Deeplink supply on MobileWorld. Task-matched deeplinks improve general VLM performance even without hybrid training. Our trained GUI-Hopper gains more from deeplinks and retains these gains when irrelevant candidates are included.}
\label{tab:abl-supply}
\centering
\fontsize{8}{9.5}\selectfont
\setlength{\tabcolsep}{2pt}
\renewcommand{\arraystretch}{1.0}
\begin{tabular*}{\linewidth}{@{\extracolsep{\fill}}ccccc@{}}
\toprule
\multirow{2}{*}{Supply} & \multicolumn{2}{c}{Qwen3.7-plus} & \multicolumn{2}{c}{GUI-Hopper (9B)} \\
\cmidrule(lr){2-3}\cmidrule(l){4-5}
 & SR (\%) $\uparrow$ & Length $\downarrow$ & SR (\%) $\uparrow$ & Length $\downarrow$ \\
\midrule
NA   & 61.5 & 25.6 & 56.7 & 23.1 \\
GT   & 62.4 & 21.4 & 63.8 & 21.7 \\
Ten  & 59.5 & 21.4 & 63.5 & 21.8 \\
Full & 53.0 & 24.1 & 62.8 & 21.7 \\
\bottomrule
\end{tabular*}
\par\smallskip
\raggedright NA: no deeplinks; GT: task-matched; Ten: retrieval top 10; Full: full catalog. Length: average steps over all trajectories.\par
\vspace{-\intextsep}
\end{wraptable}

Although deeplinks can bypass lengthy GUI navigation, agents must select task-relevant deeplinks to actually realize this efficiency. Table~\ref{tab:abl-supply} compares the effectiveness and robustness of Qwen3.7-plus and GUI-Hopper under different deeplink supply conditions.

For Qwen3.7-plus, task-matched deeplinks (GT) primarily boost navigation efficiency—reducing trajectory length by 16.25\%—even without hybrid-action training. However, expanding the catalog to Ten and Full steadily degrades its success rate (falling to 52.99\%), indicating that broader candidate sets introduce decision noise that the base model cannot filter out.

In contrast, GUI-Hopper exhibits remarkable robustness to catalog expansion, maintaining nearly identical success rates ($\sim$63\%) and step counts ($\sim$21.7) across settings of `GT', `Ten', and `Full'. Notably, Ten trails the oracle GT setting by merely 0.29 percentage points. This confirms that hybrid-action training enables the policy to reliably discern and invoke task-relevant actions from noisy pools, supporting a practical division of labor where retrieval prunes the search space while the policy executes precise selection.

\vspace{-4pt}
\section{Conclusion}
\vspace{-4pt}

In this work, we challenge the assumption that mobile agents must navigate screen-by-screen. We introduce \sysname{}, a framework that integrates app-native deeplinks as a high-level navigation channel alongside fine-grained GUI control. To enable this, we develop an automated pipeline that mines deeplinks, validates them on real devices, and grounds them in descriptions of their landing screens. Through hybrid trajectory synthesis and online reinforcement learning, the model learns a robust policy to invoke deeplinks when optimal and seamlessly fall back to GUI actions when necessary, on both synthetic environments and real mobile phones. Our evaluations on MobileWorld and MobileWorld-Real demonstrate that \sysname{} significantly improves task success and reduces trajectory lengths across all tested model scales. Notably, these gains persist even when deeplinks are withheld at inference, underscoring enhanced underlying task-planning capabilities. By bridging the gap between app-native programmatic interfaces and fine-grained visual control, this work establishes a new foundation for the next generation of mobile GUI agents.
\vspace{-2pt}
\paragraph{Weaknesses \& Limitations.}
Without system privileges, deeplink coverage is limited to screens that applications expose for external access. Furthermore, our current evaluation is restricted to the Android ecosystem. Future work will explore system-level integrations and dynamically enrich the deeplink catalog using user-specific parameters captured during live interaction.

\subsection*{AI use statement}
In this work, we used generative AI tools for drafting and polishing the text of the paper and for assisting with the implementation of experiment and analysis code. We have not used generative AI tools for generating the research ideas or the experimental design, and the remaining required-disclosure categories are not applicable to this work. Additionally, generative AI models serve as components of the method itself: as described in Section~\ref{sec:mining}, they generate and filter the natural-language descriptions in the deeplink catalog. We have reviewed all AI-assisted work: AI-drafted text was revised by the authors, and AI-assisted code was verified and tested for correctness. We take responsibility for the final content of this work, including text, claims or artifacts produced with the aid of generative AI.

\subsection*{Ethics statement}
The agent interacts with applications only through their open interfaces: GUI operations available to any user, and deeplink entry points that applications expose for external invocation by design. Our experiments did not use real users' personal information. The user-specific examples in Appendix~\ref{app:harness} illustrate potential applications of the implemented parameter-capture mechanism.

\subsection*{Reproducibility statement}
The mining pipeline is described in Section~\ref{sec:mining}, with implementation details, prompts, and statistics in Appendix~\ref{app:mining}. Appendix~\ref{app:schema} presents the deeplink schema and the prompt format seen by the model. Training settings are provided in Section~\ref{sec:setup} and Appendix~\ref{app:training}, and the harness implementation is detailed in Appendix~\ref{app:harness}. The evaluation protocol---benchmarks, task counts, supply conditions, and metrics---is specified in Section~\ref{sec:setup}. We will release the trained models and the deeplink catalog for MobileWorld.

\bibliography{references}
\bibliographystyle{iclr2027_conference}

\clearpage
\appendix
\section{Mining Pipeline: Implementation, Prompts, and Catalog Statistics}
\label{app:mining}

This appendix supplements Section~\ref{sec:mining} with implementation details, final catalog statistics, and the prompts used in the mining pipeline.

\paragraph{Static analysis.}
Candidate entries are extracted with the Shortcut-Anything tool released alongside MAS-Bench \citep{masbench}, which parses manifest declarations and shortcut definitions from the application package. Entries that differ only in how parameters are represented in the URI are consolidated by URI structure. Static extraction alone does not establish whether an entry works, whether it opens a meaningful screen, or which parameter values it accepts. The remainder of the pipeline answers these questions on real devices.

\paragraph{Exportability filtering.}
Android only allows external invocation of components that the application explicitly exports. We therefore filter out candidates whose target components are not exported.

\paragraph{Resolvability screening.}
For each remaining candidate, we use a read-only package-manager query to check whether an installed component can handle it, without launching the application or invoking a model.

\paragraph{On-device verification.}
We test the resolvable candidates on real devices and retain only those that open a meaningful screen. For parameterized entries, we test candidate values and retain a parameter only if changing its value visibly changes the landing screen.

\paragraph{Deduplication.}
Different verified entries can lead to the same screen---the search screen of one map application was reachable through 11 entries spanning 5 domain variants and 3 URI schemes. We consolidate duplicates using criteria ordered by evidential strength: repeated calls on the device, screenshot and text comparisons, and a final command-level check in which byte-identical invocation commands take precedence over the other evidence. Uncertain cases are left unmerged to avoid merging distinct entries. We also exclude entries that open transient content instances rather than stable functional screens.

\paragraph{Catalog statistics.}
The final catalog contains 1,251 agent-callable entries across 26 applications, including 246 entries with input parameters. Table~\ref{tab:app-perapp} reports the per-application distribution.

\begin{table}[ht]
\caption{Per-application counts in the verified deeplink catalog for MobileWorld-Real. The catalog contains 1,251 callable entries across 26 applications.}
\label{tab:app-perapp}
\centering
\small
\begin{tabular}{@{}lr@{\hspace{2em}}lr@{}}
\toprule
Application & Entries & Application & Entries \\
\midrule
Bilibili            & 330 & Huaxiaozhu     & 15 \\
Xiaohongshu         & 177 & Baidu Maps     & 11 \\
Weibo               & 148 & Meituan Dache  & 11 \\
NetEase Cloud Music & 128 & NetEase News   &  9 \\
Douyin              &  91 & Tencent News   &  5 \\
iQIYI               &  75 & Baidu          &  4 \\
Piaoniu             &  60 & Kuaishou       &  3 \\
AMap                &  44 & Tomato Novel   &  3 \\
Zhihu               &  43 & Tencent Video  &  2 \\
Taopiaopiao         &  35 & Baidu Tieba    &  1 \\
Didi                &  20 & Toutiao Lite   &  1 \\
Youdao              &  17 & Dongchedi      &  1 \\
Autohome            &  16 & Doubao         &  1 \\
\midrule
\multicolumn{3}{l}{\textbf{Total}} & \textbf{1,251} \\
\bottomrule
\end{tabular}
\end{table}

\paragraph{Prompts.}
A vision-language model assists with on-device verification and deduplication. During verification, the prompt guides the model to propose parameter values, invoke candidates, assess their landing screens, and write catalog descriptions in the format specified in Appendix~\ref{app:schema}. It also instructs the model to identify tracking- or session-bound parameters that should be excluded from the released catalog. This prompt is released together with the catalog. For deduplication, the two prompts below ask the model to group equivalent entries from the same application and URI base. The second also uses landing-screen screenshots as evidence. The calling program then selects which entry to retain under fixed rules.

\begin{lstlisting}[style=promptbox,title={\small\bfseries Deeplink deduplication prompt},captionpos=t]
You are deduplicating a batch of Android deeplink tools. They come from the
same app and the same URI base, but they may be entirely different features
(sharing one routing shell), or one feature inflated into several entries
by different parameter values.

Criteria (follow strictly; do not improvise):
- SAME feature = lands on the same page, and the parameters are semantically
  interchangeable (only their values differ).
- DIFFERENT features = different landing pages, or parameters that denote
  different business objects.
- Same page but resting on a different tab / sub-page -> count as different
  features, and note "same page, different tab" in the reason.
- When unsure, put the entry in its own group and mark uncertain=true.
  Do not guess: a wrong merge costs far more than a missed one.

Output strict JSON, with no explanatory text:
{"groups":[{"members":["<id>"],"reason":"...","uncertain":false}]}
Use the short ids from the input verbatim (like c01); do not rewrite or
expand them. Every input id must appear in exactly one group -- none
missing, none repeated.
Do not output a "keep" field -- which entry to keep is decided by the
caller under fixed rules, not by you.
\end{lstlisting}

\begin{lstlisting}[style=promptbox,title={\small\bfseries Deeplink deduplication prompt with screenshot evidence},captionpos=t]
Below are several deeplink candidates from the same app and the same URI
base. For each candidate you are given: id, tool name, parameter names,
textual description, and the screenshot of the landing page it actually
opened on a real device.

Judge whether they are the same feature, and split them into groups.

Criteria (follow strictly):
- SAME feature = lands on the same page, parameters only differ in value
  (e.g., different products on the same product-detail page).
- DIFFERENT features = different landing pages, or parameters that denote
  different business objects.
- Same page but resting on a different tab / sub-page -> different
  features; note it in the reason.
- A screenshot showing an error page / blank page / login page / loading
  state cannot serve as evidence of sameness -- two entries both failing
  to open does not make them the same feature. Put such an entry in its
  own group and mark uncertain=true.
- Trust the screenshot. When the screenshot conflicts with the textual
  description, follow the screenshot and state the conflict in the reason.
- When unsure, put the entry in its own group and mark uncertain=true.
  Do not guess: a wrong merge costs far more than a missed one.

Output strict JSON, with no explanatory text:
{"groups":[{"members":["<id>"],"reason":"...","uncertain":false}]}
Use the short ids from the input verbatim (like c01). Every input id must
appear in exactly one group -- none missing, none repeated.
\end{lstlisting}

\section{Deeplink Catalog Format and Agent Prompt}
\label{app:schema}

This appendix details the catalog entry format introduced in Section~\ref{sec:mining} and shows how deeplink candidates are presented to the model.

\begin{figure}[htbp]
\centering
\includegraphics[width=\linewidth]{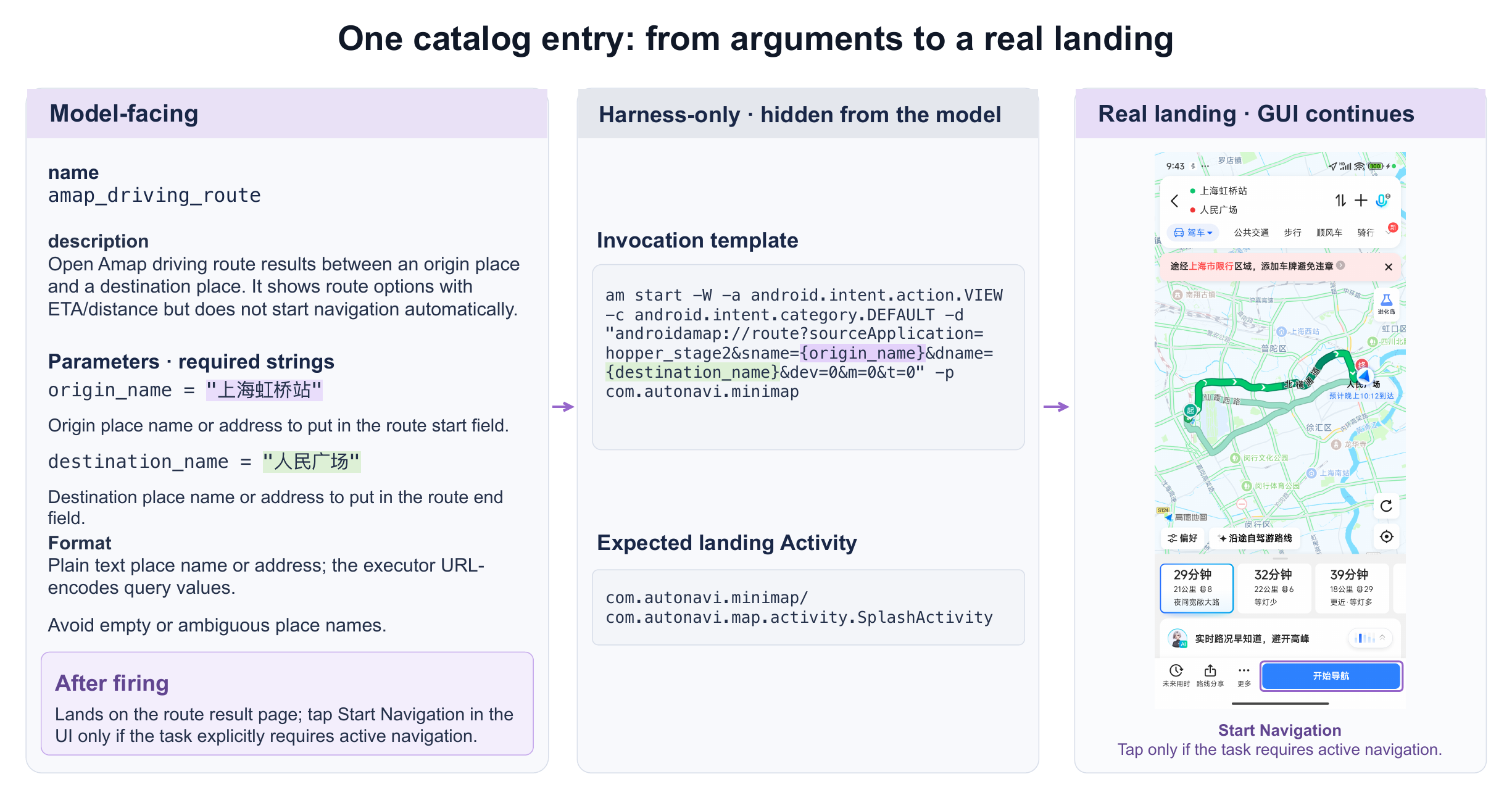}
\caption{An AMap driving-route entry connecting model-facing descriptions with harness-side execution and landing verification. The deeplink opens route results for the specified origin and destination. Starting navigation remains a GUI action when required by the task.}
\label{fig:app-schema-cases}
\end{figure}

\paragraph{Schema.}
Each catalog entry contains model-facing and harness-facing fields. The model-facing fields hold the name, the description, and the parameter definitions. The tool description names the landing screen and ends with an \texttt{After firing:} clause describing the expected outcome and any GUI actions still required. Each parameter definition specifies its meaning, format constraints, an example value, and mistakes to avoid. The harness-facing fields, hidden from the model, store a parameterized Android activity-manager command template and the expected landing Activity. The harness uses these fields to execute the call and compare the observed Activity with the expected one (Section~\ref{sec:harness}). Figure~\ref{fig:app-schema-cases} illustrates an AMap driving-route entry and its observed landing screen.

\paragraph{Prompt format.}
The model-facing fields of the selected deeplink entries are included in the system prompt. The listing below presents the system prompt for one evaluation request, with the application list omitted and the 10 candidate deeplinks retained. The prompt defines the agent role, output format, and GUI action space before presenting the deeplink candidates as function-calling schemas. A final usage guideline instructs the agent to verify task completion through the GUI rather than rely on a successful deeplink call alone.

\begin{lstlisting}[style=promptbox]
You are a GUI agent. You are given a task and your action history, with screenshots. You
    need to perform the next action to complete the task.

## Output Format
For each function call, return the thinking process in <think> </think> tags, and a json
    object with function name and arguments within <tool_call></tool_call> XML tags:
```
<think>
...
</think>
<tool_call>
{"name": "mobile_use", "arguments": <args-json-object>}
</tool_call>
```

## Action Space
{"action_type": "click", "coordinate": [x, y]}
{"action_type": "double_click", "coordinate": [x, y]}
{"action_type": "long_press", "coordinate": [x, y]}
{"action_type": "type", "text": ""}
{"action_type": "open", "text": "app_name"} # app_name must be chosen from the available
    apps listed in the Note section.
{"action_type": "drag", "start_coordinate": [x1, y1], "end_coordinate": [x2, y2]}
{"action_type": "system_button", "button": "button_name"} # Options: back, home, menu,
    enter
{"action_type": "wait", "waiting_time": "xxx"} # waiting_time denotes the wait time in
    seconds.
{"action_type": "ask_user", "text": "xxx"} # you can ask user for more information to
    complete the task.
{"action_type": "terminate", "answer": "xxx", "status": "success or failed"}

## Deeplink Tools
You are also provided with Deeplink tools, you can use them to complete the task.
{"type": "function", "function": {"name": "clock_open_timer", "description": "[Clock]
    Open Clock to the Timer screen. After firing: set duration and start timer.",
    "parameters": {"type": "object", "properties": {}, "required": []}}}
{"type": "function", "function": {"name": "calendar_open_view_or_event", "description":
    "[Calendar] Open Fossify Calendar to a date in day, month, year, agenda, or week
    view. After firing: calendar opened to requested date/view.", "parameters": {"type":
    "object", "properties": {"day_code": {"type": "string", "description": "Target date
    to open in the calendar. Format: YYYYMMDD. Example: 20261224. If omitted/empty:
    Omitted or empty uses device GMT current date. Avoid: Do not use dashed dates;
    invalid formats may crash."}, "view_to_open": {"type": "string", "description":
    "View selector: 1 month, 2 year, 3 agenda, 4 week. Format: Integer 1, 2, 3, or 4.
    Example: 1. If omitted/empty: Omit for day view; empty string is invalid. Avoid: Do
    not use empty or outside 1-4."}}, "required": []}}}
{"type": "function", "function": {"name": "clock_set_timer", "description": "[Clock]
    Create and start a countdown timer in Clock. After firing: timer starts; UI opens
    unless skip_ui=true.", "parameters": {"type": "object", "properties":
    {"length_seconds": {"type": "integer", "description": "Countdown duration in
    seconds. Format: Positive integer seconds. Example: 90. If omitted/empty: Required;
    no default."}, "message": {"type": "string", "description": "Timer label. Format:
    Plain text label. Example: Tea break. If omitted/empty: Omitted or empty uses
    default duration label."}, "skip_ui": {"type": "boolean", "description": "Whether to
    avoid foregrounding Clock. Format: Boolean true or false. Example: False. If
    omitted/empty: Omitted opens Clock; empty string is invalid. Avoid: Do not pass an
    empty boolean value."}}, "required": ["length_seconds"]}}}
{"type": "function", "function": {"name": "camera_open_photo_capture", "description":
    "[Camera] Open Android camera photo-capture UI for taking a still image. After
    firing: press shutter to capture.", "parameters": {"type": "object", "properties":
    {"output_uri": {"type": "string", "description": "Destination file URI for captured
    image. Format: file:///sdcard/.../*.jpg. Example:
    file:///sdcard/DCIM/Camera/reenrich_photo_capture.jpg. If omitted/empty: Optional;
    omit for app-managed destination. Empty is not useful. Avoid: Do not use empty or
    non-file destinations."}}, "required": []}}}
{"type": "function", "function": {"name": "mail_compose", "description": "[Mail] Open
    email compose with recipient and optional subject or body. After firing: tap Send to
    send email.", "parameters": {"type": "object", "properties": {"to": {"type":
    "string", "description": "Recipient email address. Format: RFC-style email address.
    Example: greg@example.com. If omitted/empty: Required; empty opens compose without
    recipient. Avoid: Do not use display names only."}, "subject": {"type": "string",
    "description": "Email subject line. Format: Plain text; URL-encode reserved
    characters. Example: ColdSubjectOnly. If omitted/empty: Omitted or empty leaves
    subject blank. Avoid: Do not combine with body when body must prefill."}, "body":
    {"type": "string", "description": "Email body text. Format: Plain text; URL-encode
    reserved characters. Example: ColdBodyOnly. If omitted/empty: Omitted or empty
    leaves body blank. Avoid: Subject parameter present causes body to be ignored."}},
    "required": ["to"]}}}
{"type": "function", "function": {"name": "contacts_insert", "description": "[Contacts]
    Open new-contact editor with contact fields prefilled; saving remains manual. After
    firing: tap Save to create the contact.", "parameters": {"type": "object",
    "properties": {"name": {"type": "string", "description": "Contact display name;
    editor splits full names. Format: Human name, preferably given and family name.
    Example: Alex Chen. If omitted/empty: Omitted or empty leaves name fields blank.
    Avoid: Do not expect full name to remain one field."}, "phone": {"type": "string",
    "description": "Contact phone number. Format: E.164 or local dialable number.
    Example: +14155550123. If omitted/empty: Omitted or empty leaves phone blank. Avoid:
    Phone alone does not infer a name."}, "company": {"type": "string", "description":
    "Contact organization name. Format: Plain organization name. Example: Northwind
    Labs. If omitted/empty: Omitted or empty leaves company blank."}, "email": {"type":
    "string", "description": "Contact email address. Format: Valid email address.
    Example: alex.chen@example.com. If omitted/empty: Omitted or empty leaves email
    blank."}, "job_title": {"type": "string", "description": "Role title under
    organization. Format: Plain job title. Example: Product Manager. If omitted/empty:
    Omitted or empty leaves title blank."}, "postal": {"type": "string", "description":
    "Contact postal address. Format: Single-line postal address. Example: 1 Market St,
    San Francisco, CA 94105. If omitted/empty: Omitted or empty leaves address blank."},
    "notes": {"type": "string", "description": "Contact note text. Format: Short
    free-text note. Example: Met at launch planning. If omitted/empty: Omitted or empty
    leaves notes blank."}}, "required": []}}}
{"type": "function", "function": {"name": "contacts_open", "description": "[Contacts]
    Open an existing contact's detail page by exact display name. After firing: contact
    detail page opens.", "parameters": {"type": "object", "properties": {"name":
    {"type": "string", "description": "Exact display name of an existing contact.
    Format: Exact Contacts display_name text. Example: Emily Johnson. If omitted/empty:
    Required; empty or unresolved names do not open a contact. Avoid: Do not use
    nicknames, partial names, or unknown contacts."}}, "required": ["name"]}}}
{"type": "function", "function": {"name": "calendar_insert_event", "description":
    "[Calendar] Open Calendar's new-event form with title, location, description, time,
    and all-day fields prefilled. After firing: review and tap Save to create the
    event.", "parameters": {"type": "object", "properties": {"title": {"type": "string",
    "description": "Event title. Format: Plain text. Example: DeepLink GMT Planning. If
    omitted/empty: Omitted or empty leaves title blank."}, "beginTime": {"type":
    "string", "description": "Event start time. Format: Epoch milliseconds; interpret
    date strings as GMT before firing. Example: 1760799600000. If omitted/empty: Omitted
    defaults to current time; empty long fails. Avoid: Do not local-time shift or pass
    empty strings."}, "endTime": {"type": "string", "description": "Event end time.
    Format: Epoch milliseconds; interpret date strings as GMT before firing. Example:
    1760803200000. If omitted/empty: Omitted defaults to current time; empty long fails.
    Avoid: Provide with beginTime for future events."}, "eventLocation": {"type":
    "string", "description": "Event location text. Format: Place name or address.
    Example: Conference Room 3. If omitted/empty: Omitted or empty leaves location
    blank."}, "description": {"type": "string", "description": "Event notes. Format:
    Plain text. Example: Discuss integration checkpoints. If omitted/empty: Omitted or
    empty leaves description blank."}, "allDay": {"type": "string", "description":
    "Whether the event is all-day. Format: true or false. Example: false. If
    omitted/empty: Omitted behaves as false; empty boolean fails. Avoid: Do not pass
    empty or non-boolean values."}}, "required": []}}}
{"type": "function", "function": {"name": "clock_set_alarm", "description": "[Clock]
    Create an alarm with hour, optional minute, and optional label. After firing: alarm
    created and Clock alarm list opens.", "parameters": {"type": "object", "properties":
    {"hour": {"type": "string", "description": "Alarm hour in 24-hour time. Format:
    Integer 0-23. Example: 20. If omitted/empty: Required; no default. Avoid: Do not use
    12-hour AM/PM text."}, "minutes": {"type": "string", "description": "Alarm minute.
    Format: Integer 0-59. Example: 20. If omitted/empty: Omitted defaults to 0. Avoid:
    Do not include colon-formatted time."}, "message": {"type": "string", "description":
    "Alarm label. Format: Short plain text label. Example: ReenrichAlarm. If
    omitted/empty: Omitted or empty creates an unlabeled alarm. Avoid: Avoid unquoted
    spaces in shell commands."}}, "required": ["hour"]}}}
{"type": "function", "function": {"name": "chrome_open_url", "description": "[Chrome]
    Open a complete URL in Chrome for manual reading or follow-up. After firing: page
    opened for reading.", "parameters": {"type": "object", "properties": {"url":
    {"type": "string", "description": "Complete web URL to open. Format: Absolute URL
    with scheme, usually https://. Example:
    https://example.com/?deeplink_stage2=chrome_open_url. If omitted/empty: Required;
    empty fails, omitted opens app chooser. Avoid: Do not omit scheme or pass an empty
    URL."}}, "required": ["url"]}}}

If you want to use Deeplink tools, you must output as the following format:
```
<think>
...
</think>
<tool_call>
{"name": <function-name>, "arguments": <args-json-object>}
</tool_call>
```
## Note
- Available Apps: `[... 47 application names, elided ...]`.
- You should use the `open` action to open the app as possible as you can, because it is
    the fast way to open the app.
- You must follow the Action Space strictly, and return the correct json object within
    <think> </think> and <tool_call></tool_call> XML tags.

# Deeplink shortcuts
Besides the GUI/MCP tools, the following functions are deeplink shortcuts that jump
    straight to a screen or pre-fill a form, saving navigation/typing.
- A deeplink only OPENS / PRE-FILLS. It NEVER sends, saves, or posts; finish with GUI
    actions (e.g. tap Send / Save).
- CRITICAL: a deeplink's success ONLY means the screen/form is ready, NOT that the task
    is done. Do NOT terminate or answer right after a deeplink; first finish the submit
    (tap Send / Save / Post / Confirm) and verify it on screen. Each tool description
    says whether a final submit is still needed.
- Its result is a JSON string {"success": bool, "error": ...}. Pre-filled content may be
    longer than the visible area; trust it (or scroll to verify) instead of re-typing.
- Use a shortcut only when it matches the current sub-goal; otherwise keep using the GUI
    tool.
\end{lstlisting}

\section{Training Configuration}
\label{app:training}

This appendix completes the training configuration of Section~\ref{sec:setup}. Supervised fine-tuning additionally uses a warmup ratio of 0.05, bf16 precision, and no sequence packing. At each model scale, the GUI-only and hybrid models use the same SFT configuration, differing only in training trajectories.

Online reinforcement learning initializes the policy from the SFT checkpoint. Each iteration samples 117 tasks with eight trajectories per task. We limit each trajectory to 50 interaction steps and the context to 20,480 tokens. The resulting interaction steps are used for policy updates with a global batch size of 512. The policy objective uses asymmetric clipping with $\epsilon_{\mathrm{low}}=0.2$ and $\epsilon_{\mathrm{high}}=0.28$, without KL regularization. When computing $c(\tau)$, a task-relevant deeplink receives partial credit of 0.3 if it is invoked during the trajectory but none of its invocations reaches the expected Activity.

\section{Agent Harness Implementation}
\label{app:harness}

\paragraph{Execution and feedback.}
The model generates a deeplink call in the format shown in Appendix~\ref{app:schema}. The harness fills the catalog entry's activity-manager command template with the supplied parameter values and executes the resulting command on the device to dispatch an Android Intent. Successful dispatch does not guarantee arrival at the expected screen: Android routes the Intent through the application's entry logic, which may briefly bring a splash or routing Activity to the foreground before the destination appears. The harness therefore verifies the landing by checking whether the Activity specified in the catalog entry reaches the foreground. If the target application is in the foreground but the expected Activity has not appeared, the harness waits up to 10 seconds; if the foreground package is not the target application, it stops waiting within 4 seconds. The harness then returns the outcome---whether the dispatch succeeded, which Activity was actually reached, and whether it matches the expected one. This feedback, denoted by $f_t$ in Section~\ref{sec:prelim}, appears beside the screenshot in the next observation as a text block prefixed with \texttt{Tool call result:}. If dispatch fails or the observed Activity does not match the expected one, the agent can use this feedback to continue through GUI actions.

\paragraph{Task-level retrieval.}
Retrieval runs once per task, before execution starts. A language model splits the task into navigation goals in execution order and names the application each goal belongs to. For each goal, the candidate pool is restricted to tools from the corresponding application. We use reciprocal rank fusion to combine embedding-based and BM25 rankings. The lexical scores affect the ranking but do not filter out candidates. We first select the highest-ranked candidate not already chosen from each goal's list. We then fill the remaining slots according to the fused ranking, returning up to 10 entries. To assess retrieval coverage, we manually annotated deeplink requirements for 100 tasks. Our retrieval method achieves a COMP@10 of 98.3\%, indicating high coverage of task-relevant deeplinks with a candidate set of at most 10 entries.

\paragraph{Capture in use.}
The harness captures parameter values from Intents observed during application use. These Intents can specify an action, a target component, a data URI with query parameters, and key--value extras. For a mined entry, observation provides values used in the user's own interactions---for example, an order identifier, a saved address, or a travel date---that probing alone may not recover. The user-specific examples above illustrate potential deployment scenarios. Following the filtering criteria used in mining (Appendix~\ref{app:mining}), the harness excludes parameters used for tracking or tied to a particular session. It associates retained values with the corresponding catalog entries so that later calls can reuse them.

\section{Further Analysis}
\label{app:further-analysis}

\subsection{GUI Ability without Deeplinks}\label{sec:trajectory-length}

\begin{wrapfigure}{r}{0.49\textwidth}
\vspace{\dimexpr-\intextsep-1.8\baselineskip\relax}
\vspace{-10pt}
\centering
\includegraphics[width=\linewidth,trim=0 0 0 6pt,clip]{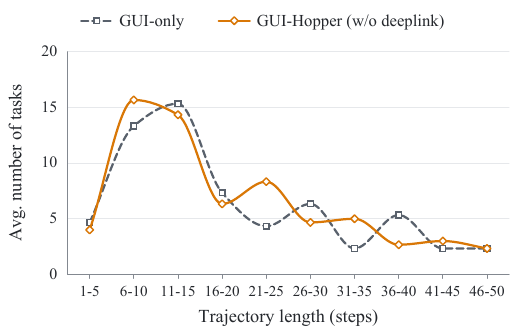}
\caption{Successful-trajectory length distributions under GUI-only evaluation on MobileWorld.}
\label{fig:trajectory-length}
\vspace{-\intextsep}
\end{wrapfigure}

Hybrid interaction should expand an agent's capabilities without making it dependent on deeplink availability. Under GUI-only evaluation, GUI-Hopper achieves higher success rates than the GUI-only trained model at all three scales (Table~\ref{tab:mw-main}). Figure~\ref{fig:trajectory-length} further compares their successful-trajectory length distributions at 9B, showing comparable execution efficiency with a slightly lower mean length for GUI-Hopper. Together, these results suggest that learning to use deeplinks complements GUI ability: the model's advantage is not confined to executions that invoke direct navigation shortcuts.

One possible explanation is that hybrid-action training encourages the model to break tasks into intermediate goals and identify the screens needed to achieve them. This planning pattern may also guide execution when only GUI actions are available. The 35B-A3B case in Figure~\ref{fig:app-case-gui-planning} illustrates this distinction. GUI-Hopper returns to the resume earlier to obtain the recipient's email address, whereas the baseline searches Mail and Contacts before returning to the document. Finding the address takes two steps for GUI-Hopper versus 21 for the baseline, while both use nine steps to compose and send the email. The savings therefore come from choosing a more direct path to the required information, rather than using fewer actions to operate the email form.

\subsection{Efficiency across Task Structures}\label{sec:gain}

\begin{table}[htbp]
\caption{Qwen3.5-9B performance by task structure on MobileWorld. Success gains are larger for single-app tasks, while trajectory reductions are larger for multi-app tasks.}
\label{tab:app-structure}
\centering
\setlength{\tabcolsep}{4pt}
\renewcommand{\arraystretch}{1.0}
\begin{tabular}{llccc}
\toprule
Task group & Model / Evaluation & SR (\%) $\uparrow$ & \multicolumn{2}{c}{Avg. length (steps) $\downarrow$} \\
 & & & Success-only & All \\
\midrule
\multirow{2}{*}{Single-app}
 & GUI-only / GUI                 & 53.3 & 16.8 & 23.7 \\
 & GUI-Hopper / Hybrid            & \textbf{68.5} & 16.5 & 21.0 \\
\midrule
\multirow{2}{*}{Multi-app}
 & GUI-only / GUI                 & 55.4 & 21.7 & 27.4 \\
 & GUI-Hopper / Hybrid            & \textbf{59.1} & \textbf{18.5} & \textbf{22.4} \\
\bottomrule
\end{tabular}
\par\smallskip
\begin{minipage}{0.88\linewidth}
\raggedright Single-app: 55 tasks; multi-app: 62 tasks; three runs per task. Success-only lengths use each condition's own successful trajectories.\par
\end{minipage}
\end{table}

Cross-application tasks require the agent to reach relevant screens in several apps, creating repeated opportunities for deeplinks to bypass GUI navigation. Table~\ref{tab:app-structure} compares the Qwen3.5-9B GUI-only model and GUI-Hopper on single-app and multi-app tasks. GUI-Hopper improves success in both groups while reducing average trajectory length over all runs by 18.4\% on multi-app tasks and 11.2\% on single-app tasks.

The 4B example in Figure~\ref{fig:app-case-compression} illustrates this cross-application use. GUI-Hopper invokes deeplinks to open Downloads and prefill an SMS, interleaving them with GUI actions to read the resume and send the message. This illustrates how app-native entry points can be combined with GUI actions to shorten multiple stages of a cross-application workflow.

\subsection{Task Execution Time}
\label{app:execution-time}

Deeplinks can save time by bypassing repeated model decisions and device interactions during GUI navigation. To assess this practical benefit, we compare the GUI-only trained model using GUI actions with GUI-Hopper using both GUI and deeplink actions. Tables~\ref{tab:mw-execution-time} and~\ref{tab:real-execution-time} report mean durations over successful and unsuccessful trajectories. For MobileWorld-Real, each comparison uses the same tasks at a given model scale and counts at most the first 50 steps of each trajectory, matching the main evaluation budget.

\begin{table}[htbp]
\caption{Recorded task durations on MobileWorld. GUI-Hopper has lower recorded times at all three scales, although the two models use different timing methods.}
\label{tab:mw-execution-time}
\centering
\small
\setlength{\tabcolsep}{5pt}
\renewcommand{\arraystretch}{1.12}
\begin{tabular*}{0.82\linewidth}{@{\extracolsep{\fill}}ccc@{}}
\toprule
Model size & GUI-only (s) $\downarrow$ & GUI-Hopper (s) $\downarrow$ \\
\midrule
4B      & 346.5 & 181.7 \\
9B      & 395.8 & 283.4 \\
35B-A3B & 493.9 & 212.9 \\
\bottomrule
\end{tabular*}
\par\smallskip
\begin{minipage}{0.82\linewidth}
\footnotesize\raggedright
\end{minipage}
\end{table}

\begin{table}[htbp]
\caption{Execution time on MobileWorld-Real. GUI-Hopper reduces mean elapsed time by 2.9--11.0\% relative to the GUI-only trained model.}
\label{tab:real-execution-time}
\centering
\small
\setlength{\tabcolsep}{5pt}
\renewcommand{\arraystretch}{1.12}
\begin{tabular*}{0.82\linewidth}{@{\extracolsep{\fill}}ccc@{}}
\toprule
Model size & GUI-only (s) $\downarrow$ & GUI-Hopper (s) $\downarrow$ \\
\midrule
4B      & 424.8 & 378.0 \\
9B      & 273.0 & 265.2 \\
35B-A3B & 271.2 & 245.4 \\
\bottomrule
\end{tabular*}
\par\smallskip
\begin{minipage}{0.82\linewidth}
\footnotesize\raggedright
\end{minipage}
\end{table}

On MobileWorld-Real, GUI-Hopper reduces mean execution time by 2.9--11.0\% across the three model scales. Its efficiency advantage therefore extends from fewer actions to less elapsed time under the evaluated configurations. MobileWorld records also show lower durations for GUI-Hopper, although their different timing methods preclude a controlled speedup estimate.

\section{Case Studies}
\label{app:failure}
\begin{figure}[htbp]
\centering
\includegraphics[width=0.9\linewidth]{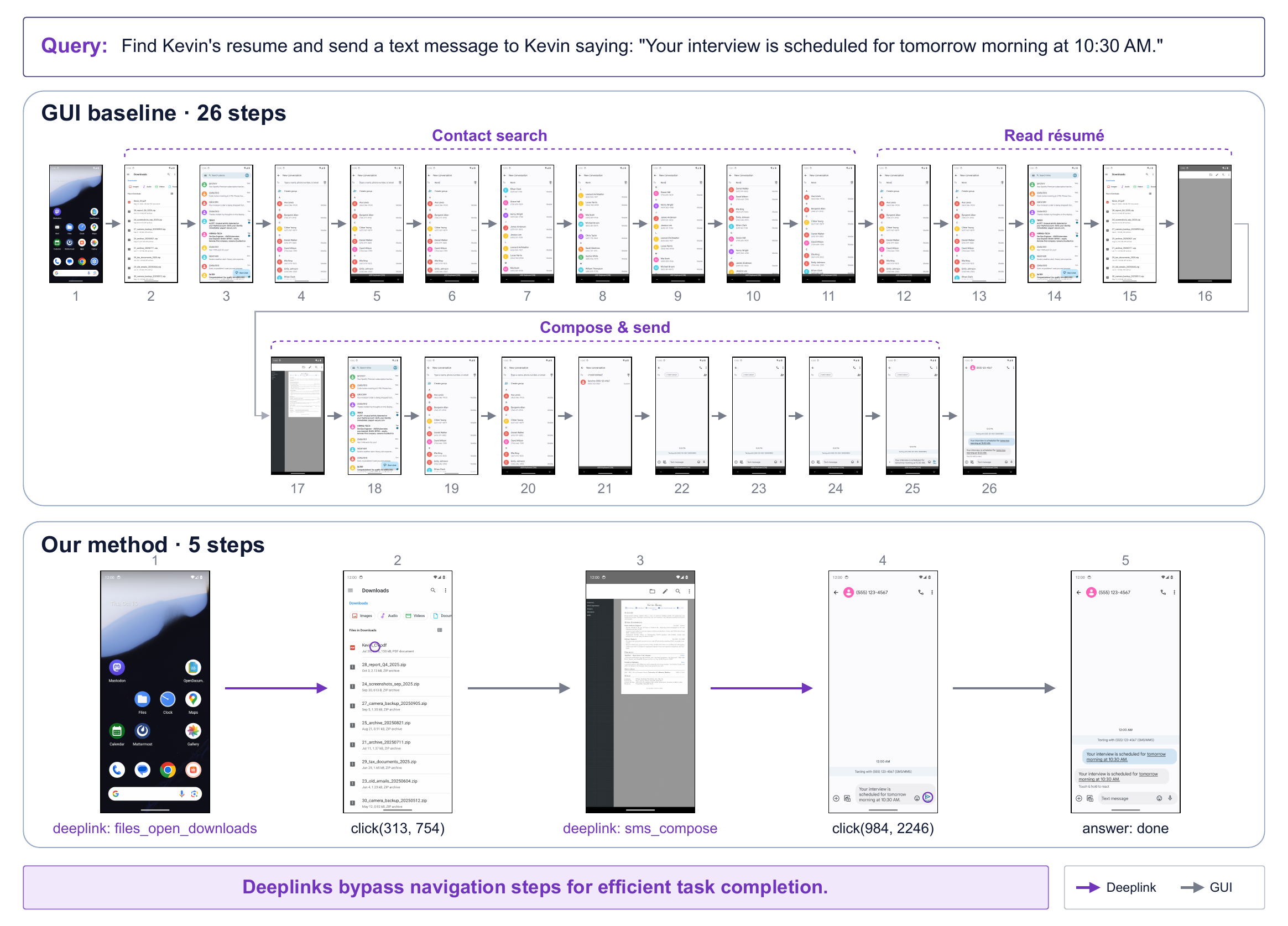}
\caption{Complete successful trajectories for an interview-invitation task on MobileWorld. GUI-Hopper takes five steps versus 26 for the GUI-only baseline. Direct navigation and SMS prefilling shorten the workflow, while GUI actions remain necessary to read the resume and send the message. Screenshots show pre-action observations, and step counts include the final answer.}
\label{fig:app-case-compression}
\end{figure}

\paragraph{Trajectory compression with deeplinks.}
Figure~\ref{fig:app-case-compression} compares two successful trajectories from the 4B models on a MobileWorld task: finding a candidate's resume and sending an interview invitation by SMS. The GUI-only baseline takes 26 steps, including contact search, repeated scrolling, and message entry. GUI-Hopper completes the task in five steps. It uses \texttt{files\_open\_downloads} to open the Downloads directory, then opens the resume through the GUI to read the recipient's phone number. It passes the number and message to \texttt{sms\_compose} and sends the prefilled message with a GUI click. The example illustrates how deeplinks shorten navigation and text entry while GUI actions complete the remaining operations.

\begin{figure}[htbp]
\centering
\includegraphics[width=0.9\linewidth]{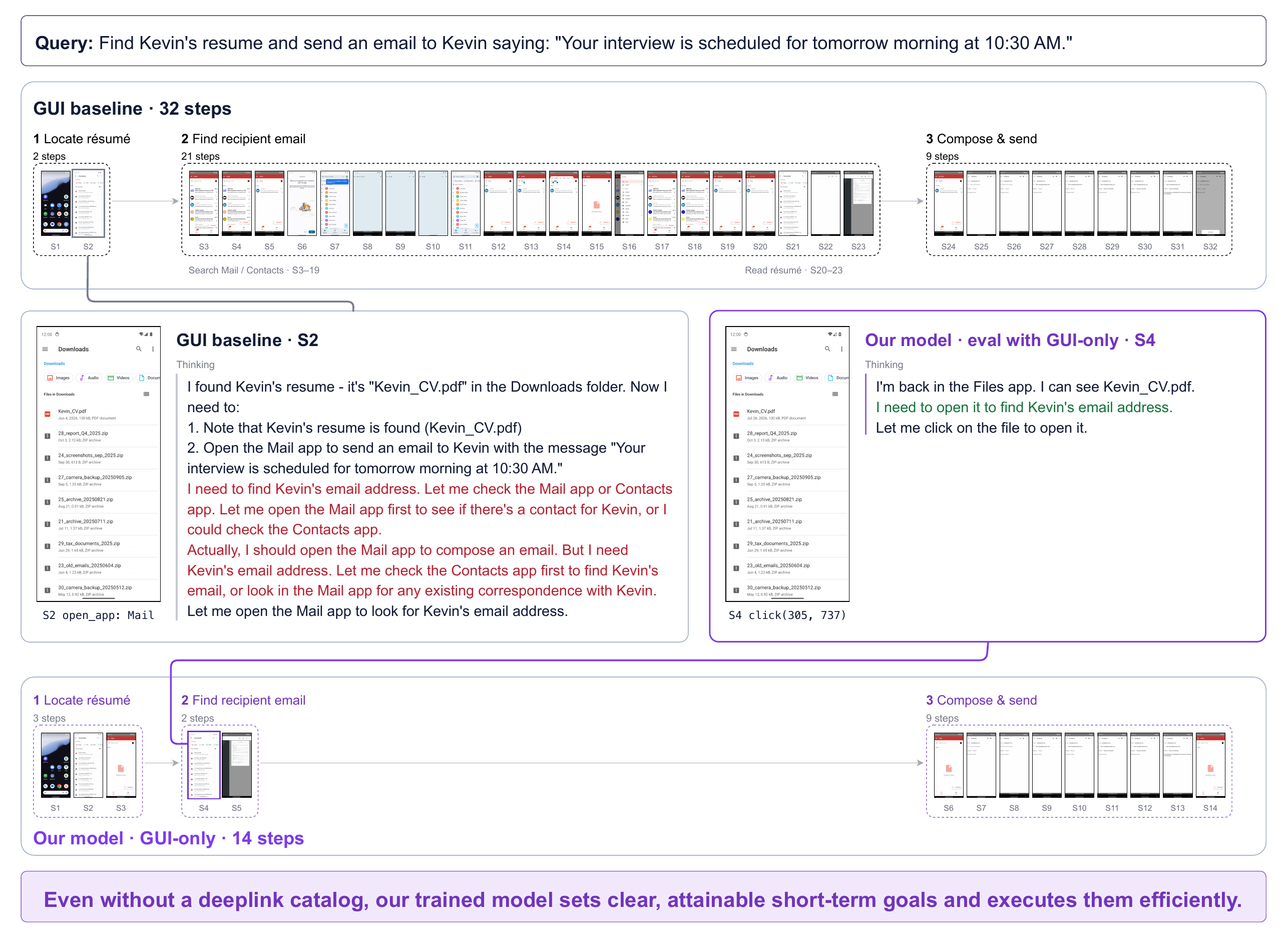}
\caption{GUI-only trajectories and model reasoning for an interview-email task on MobileWorld. Both 35B-A3B models succeed without deeplinks. GUI-Hopper obtains the recipient's address from the resume earlier, completing the task in 14 steps compared with 32 for the GUI-only baseline. Screenshots show observations before each action, and step counts include the final answer.}
\label{fig:app-case-gui-planning}
\end{figure}

\paragraph{Task planning without deeplinks.}
Figure~\ref{fig:app-case-gui-planning} examines a different source of trajectory compression: choosing where to obtain the information needed for the next operation. Both 35B-A3B models complete the interview-email task without deeplinks, but the GUI-only baseline takes 32 steps compared with 14 for GUI-Hopper. After locating the resume, the baseline searches Mail and Contacts before returning to read it. GUI-Hopper also briefly opens Mail, but returns to the resume earlier to obtain the recipient's email address. In the illustrated stage breakdown, finding the address takes 21 steps for the baseline and two for GUI-Hopper, while composing and sending the email takes nine steps for each. The main difference is therefore how the models obtain the required information, rather than how they operate the email form. This case illustrates how identifying a useful intermediate goal can reduce unnecessary exploration even when execution relies on GUI actions.

\begin{figure}[htbp]
\centering
\includegraphics[width=0.9\linewidth]{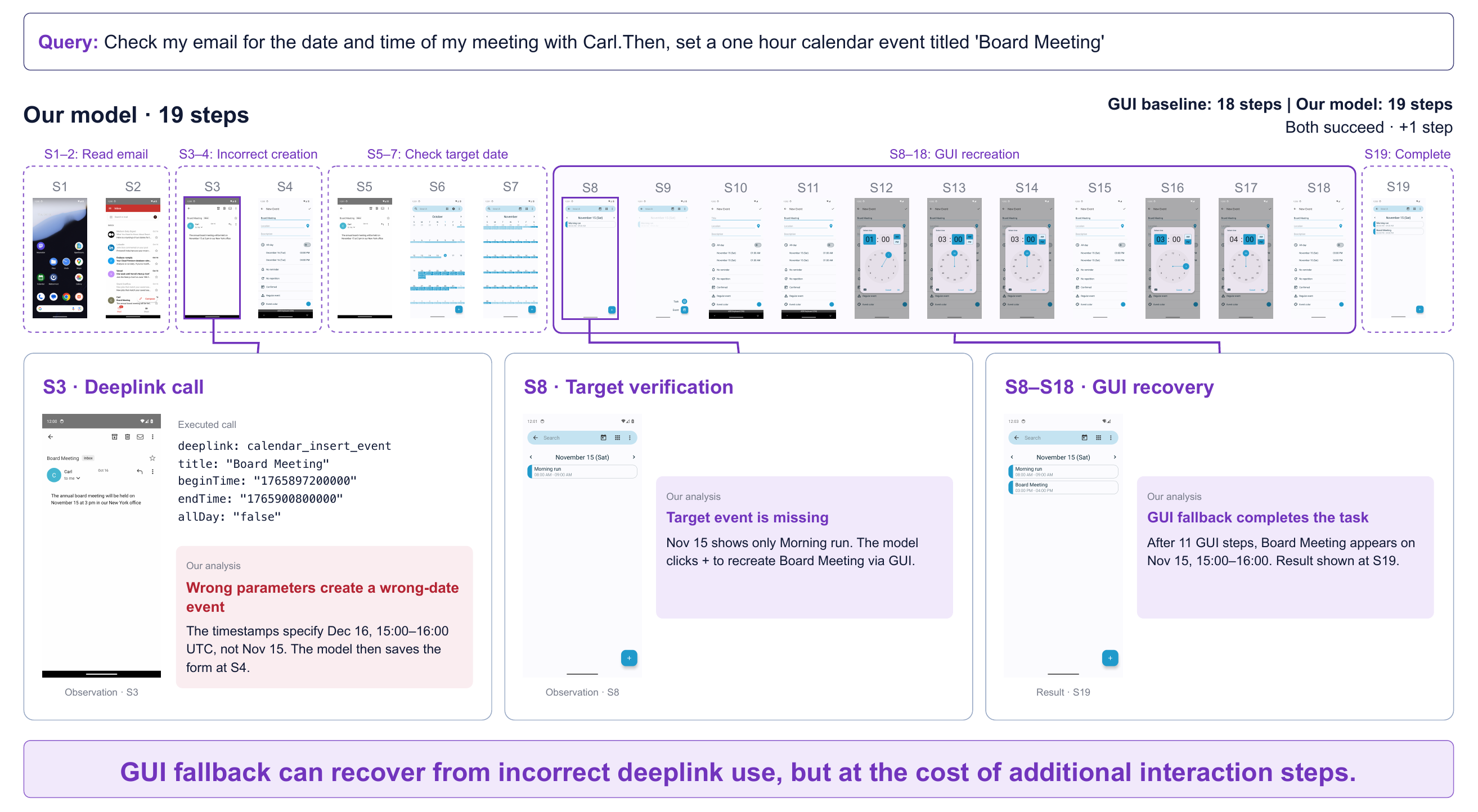}
\caption{GUI recovery after incorrect deeplink use on a MobileWorld calendar task. The call reaches the expected Activity but prefills the wrong date, showing why the model must also check task-relevant screen contents. GUI-Hopper later recreates the event through GUI actions, taking 19 steps versus 18 for the GUI-only baseline. Removal of the earlier wrong-date event is not shown.}
\label{fig:app-case-gui-recovery}
\end{figure}

\paragraph{GUI recovery after incorrect deeplink use.}
Figure~\ref{fig:app-case-gui-recovery} illustrates an error that landing verification alone cannot detect. Asked to schedule a meeting on November 15, the 9B GUI-Hopper supplies timestamps for December 16 to \texttt{calendar\_insert\_event}. The call passes landing verification because it reaches the expected Calendar Activity. However, the model overlooks the incorrect date shown in the form and saves the event. When it checks November 15, it notices that the meeting is missing and recreates it through GUI actions. The recreation takes 11 steps within a 19-step trajectory, compared with 18 steps for the successful GUI-only baseline. Reaching the expected screen therefore does not establish that the supplied parameters satisfy the task: the model must also check the screen's task-relevant contents. Here, GUI recovery achieves the requested outcome at an interaction cost, although the trajectory does not show removal of the earlier wrong-date event.

\end{document}